\documentclass[runningheads]{llncs}
\usepackage{amsmath}
\usepackage{amssymb}
\usepackage{mathtools}
\usepackage{booktabs}
\usepackage[depth=subsection]{bookmark}
\usepackage{graphicx}
\usepackage{xcolor}
\usepackage{hyperref}
\newcommand\given[1][]{\:#1\vert\:}

\begin{document}
\title{Target Robust Discriminant Analysis}
\author{Wouter M. Kouw\inst{1} \and
Marco Loog\inst{2,3}}
\authorrunning{W.M. Kouw \& M. Loog}
\institute{TU Eindhoven, Groene Loper 19, Eindhoven, the Netherlands \and
TU Delft, Van Mourik Broekmanweg 6, Delft, the Netherlands 
\and University of Copenhagen, Universitetsparken 1, Copenhagen, Denmark
}
\maketitle              % typeset the header of the contribution
\begin{abstract}
In practice, the data distribution at test time often differs, to a smaller or larger extent, from that of the original training data. Consequentially, the so-called source classifier, trained on the available labelled data, deteriorates on the test, or \emph{target}, data.  Domain adaptive classifiers aim to combat this problem, but typically assume some particular form of domain shift. Most are not robust to violations of domain shift assumptions and may even perform worse than their non-adaptive counterparts. We construct robust parameter estimators for discriminant analysis that \emph{guarantee} performance improvements of the adaptive classifier over the non-adaptive source classifier.
\keywords{Domain Adaptation \and Robustness \and Discriminant Analysis}
\end{abstract}
\section{Introduction} \label{intro}
Domain adaptation is a supervised learning setting where labelled training data is drawn from one distribution (\emph{source domain}) and unlabelled test data is drawn from another distribution (\emph{target domain}) \cite{ben2010theory,kouw2019review}. Often, adapting a source domain classifier, i.e., changing predictions to suit the target domain, is the only means by which one can potentially obtain satisfactory performance. 
Unfortunately, many domain adaptive classifiers assume some relationship between the domains, such as that only the covariates have shifted between domains and not the posterior distributions. They are not robust to violations of such assumptions and can subsequently perform \emph{worse} than non-adaptive classifiers.

%\subsection{Contributions}
\paragraph{}
We formulate a conservative adaptive classifier that always performs at least as well as the non-adaptive one. More specifically, a core contribution of this paper is that we construct estimators that produce estimates with an empirical target risk that is \emph{always} smaller or equal to the target risk of the source classifier. Since only the performance of the given target samples is considered, our result is transductive in nature \cite{vapnik1998statistical}. 
 Importantly, our guarantees are obtained without making \emph{any} domain shift assumptions such as covariate shift or the existence of a domain-invariant subspace \cite{kouw2019review}.  Furthermore, we show that in the case of classical likelihood-based discriminant analyses \cite{mclachlan2004discriminant}, the estimator will produce \emph{strictly} smaller risks (i.e. larger log-likelihoods) \emph{almost surely}, i.e., with probability 1. To the best of our knowledge, this is the first demonstration of a performance guarantee for a target classifier compared to the source classifier.

%---------------
%---------------
%---------------
%---------------
%---------------

\section{Robust Target Domain Estimator}\label{trr}
Consider a feature space $\mathcal{ X} \subseteq \mathbb{R}^{D}$ and class labels $\mathcal{ Y} = \{1, \dots, K\}$. Let $\mathcal{ S}$ denote the source domain, with $n$ samples drawn from the source domain's joint distribution, $p_{\mathcal{ S}}(x,y)$, collected as the set $\{(x_{i},y_{i})\}_{i=1}^{n}$. Similarly, let $\mathcal{ T}$ denote the target domain, with $m$ samples drawn from the target domain's joint distribution, $p_{\mathcal{ T}}(x,y)$, collected as $\{(z_{j},u_{j})\}_{j=1}^{m}$. The goal is to predict the unknown target labels $u$ (transductive setting), using only the unlabelled target samples $\{z_j\}_{j=1}^{m}$ and the labelled source samples $\{(x_i,y_i)\}_{i=1}^{n}$.

%\subsection{Target Risk}
\paragraph{}
The empirical risk in the source domain is defined as the average loss $\ell$ of a classification function $h$ over the source samples: $\hat{R} \left(h \given x,y \right) \! \triangleq \! \frac{1}{n} \sum_{i=1}^{n} \ell \left(h \given x_i, y_i \right)$. The \emph{source classifier} is the classifier that minimizes the empirical source risk:
\begin{align}
\hat{h}^{\mathcal{ S}} \triangleq \arg \min_{h \in \mathcal{H}} \ \hat{R} \left( h \given x, y \right) \, ,
\end{align}
where $\mathcal{H}$ refers to the hypothesis space. Evaluating the source classifier is typically done through the classification error. Arguably, a more appropriate evaluation is to consider the risk itself, given that it is the surrogate loss that is being optimized \cite{loog2016measuring}.
We evaluate $\hat{h}^{\mathcal{ S}}$ based on the empirical target risk:
\begin{align}
	\hat{R} \big( \hat{h}^{\mathcal{ S}} \given z,u \big) = \frac{1}{m} \sum_{j=1}^{m} \ \ell \big( \hat{h}^{\mathcal{ S}} \given z_j, u_j \big) \, . \label{hs_RT}
\end{align} 

In our objective function, the source classifier's target risk in \eqref{hs_RT} is subtracted from the target risk of a prospective target classifier $h$:
\begin{align}
	\hat{R} \big( h \given z,u \big) \ - \ \hat{R} \big( \hat{h}^{\mathcal{ S}} \given z,u \big) \label{contrast} \, .
\end{align}
The target risk of the source classifier will act as a bound on the hypothesis space during minimization:
\begin{lemma} \label{lemma1}
For fixed samples $z$ and labels $u$, the difference in empirical target risks between a classifier $h \in \mathcal{H}$ and $h^{\cal S} \in \mathcal{H}$ is less than or equal to $0$:
\begin{align}
   \underset{h \in \mathcal{H}}{\min} \ \hat{R} \big(h \given z,u \big) \ - \ \hat{R} \big( \hat{h}^{\mathcal{ S}} \given z,u \big) \leq 0 \, .
\end{align}
\end{lemma}
\begin{proof}
Let $\tilde{h} \triangleq \min_{h \in \mathcal{H}} \hat{R} \big( h \given z,u \big) \ - \ \hat{R} \big( \hat{h}^{\mathcal{ S}} \given z,u \big)$. Since $\tilde{h}$ and $\hat{h}^{\mathcal{S}}$ are elements of the same hypothesis space, $\tilde{h} = h^{\mathcal{S}}$ is a potential solution to the minimization problem. In that case, the difference in target risks would be $0$. The estimate $\hat{h}^{\cal S}$ can always be recovered, which implies all $h \in \mathcal{H}$ that would lead to risk differences greater than $0$ are not valid minimizers. \qed
\end{proof}

Equation \ref{contrast} still contains the unknown target labels $u$. To be able to guarantee a better or equal performance to that of the source classifier, we use a worst-case labelling, achieved by introducing a hypothetical labelling $q$ and \emph{maximizing} the difference in risks: $\hat{R} \left(h \given z, u \right) \leq \max_q \ \hat{R} \left( h \given z, q \right)$.

Note that for any classifier $h$, the risk with respect to this worst-case labelling will always be larger than the risk with respect to the true target labelling.
%\subsection{Final Robust Objective}
Combining the difference in risks from Equation \ref{contrast} with the hypothetical labelling $q$ results in the following risk function: 
\begin{align}
	\hat{R}^{\mathcal{T}} \big(h \given \hat{h}^{\mathcal{ S}}, z, q \big) \triangleq \hat{R}(h \given z, q) - \hat{R}(\hat{h}^{\cal S} \given z, q) \label{tr} \, .
\end{align}
We refer to the risk in Equation \ref{tr} as the Target Robust (TR) risk. Minimizing it with respect to a classifier $h$ and maximizing it with respect to the hypothetical labelling $q$, leads to a TR classifier:
\begin{align}
	\hat{h}^{\mathcal{ T}} \triangleq \arg \underset{h \in \mathcal{H}}{\min} \ \underset{q \in \mathcal{Y}^m}{\max} \ \hat{R}^{\mathcal{T}} \big( h \given \hat{h}^{\mathcal{ S}},z,q \big) \, . \label{h_T}
\end{align}
The TR risk only considers the given target samples $\{z_j\}_{j=1}^{m}$ and is, therefore, a transductive approach \cite{vapnik1998statistical}. 

%\subsection{Label Relaxation}
The maximization over a set of discrete labels is a difficult combinatorial problem. Therefore, we apply a relaxation and represent the hypothetical labelling probabilistically, $q_{jk} := p(u_j =k\given z_{j})$. That is, $q_j$ is a non-negative vector of $K$ elements that sum to $1$. As such, it represents an element of the standard $K-1$ simplex $\Delta_{K-1}$. For $m$ samples, an $m$-dimensional $K-1$ simplex $\Delta_{K-1}^{m}$ is taken.  This means that the loss for every element $z_j$ becomes a weighted sum,
\begin{equation}\label{sumloss}
\ell \big(h \given z_j, q_j \big) = \sum_{k \in \mathcal{Y}} q_{jk} \, \ell \big(h \given z_j, k \big) \, ,
\end{equation}
and the maximization in Equation \ref{h_T} will be over $q \in \Delta_{K-1}^m $ instead of $q \in \mathcal{Y}^m$.
Known, deterministic labels can of course also be represented probabilistically, for example $ y_i =1 \ \Leftrightarrow \ p( y_i = 1 \given x_i) = 1$ and $p( y_i \neq 1 \given x_i) = 0$. Hence, in practice, both $y_i$ and $u_j$ can be represented as $1 \times K$-vectors with the $k$-th element marking the probability that sample $i$ or $j$ belongs to class $k$ (a.k.a. a one-hot encoding). 

%\subsection{Robustness Lemma}
\paragraph{}
With the relaxation from "hard'' to "soft'' labels, we can say the following:
\begin{lemma} \label{truelemma}
For fixed samples $z$ and source classifier $\hat{h}^{\mathcal S}$, the Target Robust risk will be lower or equal to $0$ at its saddle point with respect to both $h$ and $q$:
\begin{align}
   \underset{h \in \mathcal{H}}{\min} \ \underset{q \in \Delta_{K-1}^m}{\max} \hat{R}^{\mathcal{T}} \big( h \given \hat{h}^{\mathcal{ S}},z,q \big) \leq 0 \, .
\end{align}
\end{lemma}

\begin{proof}
In the given minimax problem, we first go over all $h \in \mathcal{H}$ and find, for every $h$, the $q_h \in \Delta_{K-1}^m$ that maximizes the Target Robust risk. In a second step, we take the minimum of $\hat{R}^{\mathcal{T}} \big( h \given \hat{h}^{\mathcal{ S}},z,q_h \big)$ over all $h$.  Given that $q_h$ is fixed, Lemma \ref{lemma1} applies, which means the resulting Target Robust risk will be less than or equal to 0. \qed
\end{proof}

%------------------------------
%------------------------------
%------------------------------
%------------------------------
%------------------------------

\section{Discriminant Analyses} \label{case_da}
Lemma \ref{truelemma} tells us that our target classifier performs at least as well as the source classifier on the given target samples. For classical linear and quadratic discriminant analysis, we are able to show that strict improvements are obtained \emph{almost surely}, i.e., with probability 1.

%\subsection{Model}
\paragraph{}
In discriminant analysis, the data from each class is modelled with a Gaussian distribution, proportional to the class prior \cite{mclachlan2004discriminant}. We maintain a parameter vector $\theta_k$ for each class, consisting of the prior, mean and covariance matrix; $\theta_k = (\pi_k, \mu_k, \Sigma_k)$. One obtains an empirical risk minimization formulation by taking the negative log-likelihoods as the loss function: $\ell(\theta \given x, y) = \sum_{k=1}^{K} - y_{k} \log \big[ \pi_k \ \mathcal{N}(x \given \mu_k, \Sigma_k) \big] $. Note the resemblance to Equation \ref{sumloss}.

If each class is modelled with a separate covariance matrix, the resulting classifier is known as \emph{quadratic discriminant analysis} (QDA) \cite{mclachlan2004discriminant}. For target data $z$ and probabilistic labels $q$, the risk is formulated as:
\begin{align}
\hat{R}_{\text{QDA}} (\theta \given z, q) \triangleq \frac{1}{m} \sum_{j=1}^{m} \sum_{k=1}^{K} - q_{jk} \log \big[ \pi_k \ \mathcal{N}(z_{j} \given \mu_k, \Sigma_k) \big] \label{qda} \ \, . 
\end{align}
Note that the risk is now expressed in terms of classifier parameters $\theta$, as opposed to the classifier $h$. Plugging the risk from \eqref{qda} into \eqref{tr}, the full {\sc TR-QDA} risk becomes:
\begin{align}
	\hat{R}_{\text{QDA}}^{\mathcal{T}}(\theta \given \ \hat{\theta}^{\mathcal{ S}}, z, q) \triangleq&\ \hat{R}_{\text{QDA}} (\theta \given z, q) - \hat{R}_{\text{QDA}} (\hat{\theta}^{\cal S} \given z, q) \label{TR-qda} \, \, , 
\end{align}
where the estimate itself is:
\begin{align}
    \hat{\theta}^{\mathcal{ T}} \triangleq \underset{\theta \in \Theta}{\arg \min} \underset{q \in \Delta^{m}_{K-1}}{\max} \hat{R}_{\text{QDA}}^{\mathcal{T}} (\theta \given \hat{\theta}^{\mathcal{ S}}, z, q) \, .
\end{align}

%\subsection{Linear Discriminant Analysis}
If the model is constrained to share a single covariance matrix for each class, the resulting classifier is a linear function of the feature values and hence is termed \emph{linear discriminant analysis} (LDA).  The optimal overall class-covariance matrix $\Sigma$ can be determined with $\Sigma = \sum_{k=1}^{K} \pi_k \Sigma_k$.

\subsection{Performance Improvement Guarantee}
Discriminant analysis has a special property: it obtains a \emph{strictly} smaller risk. In other words, this parameter estimator is \emph{guaranteed to improve its performance} - on the given target samples, and in terms of risk - over the source classifier.\\ 

\begin{theorem} \label{thm:trda}
Let the number of target samples from a continuous target distribution be greater than its number of features. The empirical discriminant analysis risk $\hat{R}_{\text{DA}}$, i.e., the negative log-likelihood over the target samples, of the TR estimated parameters $\hat{\theta}^{\mathcal{ T}}$ is almost surely strictly smaller than for the source parameters $\hat{\theta}^{\mathcal{ S}}$.  In other words, with probability one we have the strict inequality
\[
	\hat{R}_{\text{DA}} \big( \hat{\theta}^{\mathcal{ T}} \given z,u \big) \ < \ \hat{R}_{\text{DA}} \big( \hat{\theta}^{\mathcal{ S}} \given z,u \big) \, .
\]
\end{theorem}

\begin{proof}
Let $\{(x_i,y_i)\}_{i=1}^{n}$ be a data set of size $n$ drawn \emph{i.i.d.} from a continuous source distribution defined over feature space $\mathcal{ X} \subseteq \mathbb{R}^{D}$ and label space $\mathcal{ Y} = \big\{ y = \{0, 1\}^{K} : \sum_{k=1} y_k =1 \big\}$. Similarly, let $\{(z_j,u_j)\}_{j=1}^{m}$ be a data set of size $m>D$, drawn \emph{i.i.d.} from a continuous target distribution defined over ${\cal X} \times {\cal Y}$. Consider a discriminant analysis model parametrized by $\theta = (\pi_1, .., \pi_K, \mu_1, .., \mu_K, \Sigma_1, .. \Sigma_K)$ with empirical risk defined as
\begin{align}
\hat{R}_{\text{QDA}}(\theta \given x, y) =&\ \frac{1}{m} \sum_{j=1}^{m} \sum_{k=1}^{K} - y_{ik} \log [\pi_k \ \mathcal{N}(x_i \given \mu_k, \Sigma_k) ] \ \, .
\end{align}
Let $\hat{\theta}^{\mathcal{ S}}$ be the parameters estimated on labelled source data
\begin{align}
	\hat{\theta}^{\mathcal{ S}} = \underset{\theta \in \Theta}{\arg \min} \ \hat{R}_{\text{QDA}} \big( \theta \given x, y \big)
\end{align}	
and let $(\hat{\theta}^{\cal T},q^{*})$ be the parameters and worst-case labelling estimated by mini-maximizing the Target Robust risk:
\begin{align}
	\hat{\theta}^{\cal T},\, q^{*} = \underset{\theta \in \Theta}{\arg \min}\ \underset{q \in \Delta_{K-1}^{m}}{\arg \max}\ \hat{R}_{\text{QDA}} \big( \theta \given z,q \big) - \hat{R}_{\text{QDA}} \big( \hat{\theta}^{\mathcal{ S}} \given z,q \big) \, .
\end{align}
Lemma \ref{truelemma} tells us that 
\begin{align}
	\hat{R}_{\text{QDA}} \big( \hat{\theta}^{\cal T} \given z,q^{*} \big) - \hat{R}_{\text{QDA}} \big( \hat{\theta}^{\mathcal{ S}} \given z,q^{*} \big) \leq 0  \, . \label{nonpos2}
\end{align}
Since this holds for the worst-case labelling $q^{*}$, it must also hold for the true labelling $u$:
\begin{align}	
	\hat{R}_{\text{QDA}} \big( \hat{\theta}^{\mathcal{ T}} \given z,u \big) \ \leq \ \hat{R}_{\text{QDA}} \big( \hat{\theta}^{\mathcal{ S}} \given z, u \big) \label{crisks} \, .
\end{align}
The Equality in \eqref{crisks} occurs with probability $0$, which can be shown as follows. Firstly, note that the total mean for the source classifier consists of the weighted combination of the class means, resulting in the overall source sample average
\begin{align}
	\hat{\mu}^{\mathcal{ S}} =& \ \sum_{k=1}^{K} \hat{\pi}^{\mathcal{ S}}_k \ \hat{\mu}^{\mathcal{ S}}_k  \
	= \sum_{k=1}^{K} \frac{\sum_{i=1}^{n} y_{ik}}{n} \left[ \frac{1}{\sum_{i=1}^{n} y_{ik}} \sum_{i=1}^{n} y_{ik} x_i \right] \
	= \frac{1}{n} \sum_{i=1}^{n} x_i \label{source_totmean} \, .
\end{align}
The total mean for the TP-QDA estimator is similarly defined, resulting in the overall target sample average:
\begin{align}
	\hat{\mu}^{\mathcal{ T}} \! \! = \! \sum_{k=1}^{K} \hat{\pi}^{\mathcal{ T}}_{k} \hat{\mu}^{\mathcal{ T}}_k
	\! = \! \sum_{k=1}^{K} \frac{\sum_{j=1}^{m} \! q^{*}_{jk}}{m} \! \left[ \! \frac{1}{\sum_{j=1}^{m} q^{*}_{jk}} \sum_{j=1}^{m} q^{*}_{jk} z_j \right] \!
	\! = \! \sum_{k=1}^{K} \! \frac{1}{m} \! \sum_{j=1}^{m} q^{*}_{jk} z_j 
	\! = \! \frac{1}{m} \! \sum_{j=1}^{m} z_j \label{tr_totmean2} \, .
\end{align}
Because $q^{*}$ consists of probabilities, the sum over classes $\sum_{k=1}^{K} q^{*}_{jk}$ in Equation \ref{tr_totmean2} is $1$, for every sample $j$.

Secondly, the TR objective function is quasi-convex-concave. In fact, it is linear in terms of $q$. Since its domain $\Delta_{K-1}^m$ is compact, Sion's theorem holds which allows for interchanging the order of the minimization and the maximization \cite{sion58}.  This implies %(see \cite{dresher61}) 
that the minimax solution is a saddle point and that the optimal parameter estimates $\hat{\theta}^{\cal T}$ for the discriminant analysis are unique, because the objective function is strictly (quasi-)convex in terms of these parameters when $m>D$ \cite{mclachlan2004finite}.

Now, equal risks for the source and target parameter sets on the worst-case labelling $q^*$, i.e., $\hat{R}_{\text{QDA}} \big( \hat{\theta}^{\mathcal{ T}} \given z,q^* \big) = \hat{R}_{\text{QDA}} (\hat{\theta}^{\mathcal{ S}} \given z,q^*)$, implies equality of the total means, $\hat{\mu}^{\mathcal{ T}}$ = $\hat{\mu}^{\mathcal{ S}}$, because $\hat{\theta}^{\cal T}$ is the unique minimizer of a strictly convex risk. 
By Equations \ref{source_totmean} and \ref{tr_totmean2}, equal total means implies equal sample averages: $\frac{1}{m} \sum_{j=1}^{m} z_j = \frac{1}{n} \sum_{i=1}^{n} x_i$. Given a set of source samples, drawing a set of target samples such that its average is \emph{exactly equal} to the average of the source samples, is an event that has probability $0$ under continuous distributions. Therefore, a strictly smaller risk occurs almost surely. 
% for the worst-case labelling $q^*$ and therefore also occurs for the true labelling $u$.  
In other words, with probability 1, we have that
\begin{align}
	\hat{R}_{\text{QDA}} \big( \hat{\theta}^{\mathcal{ T}} \given z,u \big) \ < \ \hat{R}_{\text{QDA}} \big( \hat{\theta}^{\mathcal{ S}} \given z,u \big) \, .
\end{align}
This concludes the proof for the case of QDA. The proof for LDA follows from plugging in $\Sigma$ for $\Sigma_k$. Since this does not alter the mean estimators, the Equality in \eqref{crisks} still occurs with probability $0$. \qed
\end{proof}

\subsection{Optimization}
As pointed out in the proof of Theorem \ref{thm:trda}, we seek a saddle point to a quasi-convex-linear problem. That can be found by first performing a gradient descent step with respect to $h$ (or, equivalently, $\theta$), followed by a gradient ascent step with respect to $q$. For discriminant analyses models, the minimization with respect to $\theta$ has a closed-form solution:
\begin{align}
	\pi_{k} =&\ \frac{1}{m} \sum_{j=1}^{m} q_{jk} \, , \qquad 
	\mu_{k} = \big( \sum_{j=1}^{m} q_{jk} \big)^{-1} \sum_{j=1}^{m} q_{jk} z_{j}  \, ,   \nonumber\\
	\Sigma_k =&\ \big( \sum_{j=1}^{m} q_{jk} \big)^{-1} \sum_{j=1}^{m} q_{jk}(z_{j} - \mu_{k}) (z_{j} - \mu_{k})^{\top} \, .  
\end{align}
One encounters the same solutions in the M step of EM-based Gaussian mixture modelling, where data points also have probabilistic class assignments \cite{mclachlan2004finite}. To ensure the updated $q$ remains on the simplex, it is projected back after each gradient step. The projection $\mathcal{ P}$ maps a point outside the simplex $a$ to the point $b$ on the simplex that is closest in terms of Euclidean distance: $\mathcal{P}(a) = \arg \min_{b \in \Delta} \| a - b \|_2$ \cite{condat2016fast}. 
The projection complicates the computation of the step size, which we replace by a learning rate $\alpha^t$ decreasing over iterations $t$. This results in the overall update: $q^{t+1} \leftarrow \mathcal{ P}(q^{t} + \alpha^{t} \nabla q^{t})$.

A gradient descent-gradient ascent procedure for globally convex-linear objectives is guaranteed to converge to a saddle point (c.f. Proposition 4.4 and Corollary 4.5 in \cite{cherukuri2017saddle}).

%------------------------------
%------------------------------
%------------------------------
%------------------------------
%------------------------------

\section{Experiments}\label{exp_gen}
Our contribution is first and foremost theoretical. Nevertheless, we perform an experiment on a natural data set comparing the empirical target risks of our TR classifiers with source classifiers (S-LDA and S-QDA) as well as classifiers trained on labelled target data (T-LDA and T-QDA), which represent the best possible performance of the models. Furthermore, we perform an experiment comparing our estimator to other domain-adaptive classifiers. Since these do not incorporate the same loss as the DA models, we measure performance in area under the ROC-curve (AUC).

The data set we used is split geographically into domains. The goal is to predict heart disease in patients from 4 different hospitals \cite{Dua2019}. These are located in Hungary, Switzerland, California and Ohio. Each hospital can be considered a domain because patients are measured on the same biometrics but the local patient populations differ. For example, the age distributions are shifted between countries. The data set was pre-processed using z-scoring.

We compared to Kernel Mean Matching (KMM) \cite{huang2007correcting}, Robust Covariate Shift Adjustment (RCSA) \cite{wen2014robust}, the Robust Bias-Aware (RBA) classifier  \cite{liu2014robust} and Transfer Component Analysis (TCA) \cite{pan2011domain}. KMM represents a standard importance-weighted classifier, which assumes covariate shift between domains. RBA and RCSA still assume covariate shift, but incorporate robust importance weight estimators. TCA represents an alternative domain shift assumption, namely the existence of a feature subspace common to both domains. These methods are discussed further in the Related Work section (Sec.~\ref{sect:rel}). All methods were trained with both a logistic and quadratic loss, and the better performing loss was chosen. For RCSA, we used the authors' implementation, which incorporates a support vector machine with Gaussian kernel. All methods use $L^2$-regularization. Since no labelled target data is available for validation, the regularization parameter was set to $0.01$ for logistic and $0.01 n$ for quadratic losses.

\begin{table}[htb]  
\caption{Target risks (average negative log-likelihoods) for all pairwise combinations of domains in heart disease data set (O='Ohio', C='California', H='Hungary' and S='Switzerland'). Smaller values are better.}    
\label{tab:hdis_tr}
%\setlength{\tabcolsep}{1pt}
%\centering
\setlength{\tabcolsep}{8.5pt}
\renewcommand{\arraystretch}{0.9}
\begin{tabular}{ r l | c c  c | c c c }
$\mathcal{S}$ & $\mathcal{T}$ &  S-LDA & TR-LDA & T-LDA &  S-QDA & TR-QDA & T-QDA \\
\midrule
O & H &  -53.55 	& -57.18 	& -57.35 	&  -53.55 	& -57.20 	& -57.62 \\
O & S &  -8.293 	& -16.76 	& -17.54 	&  -8.293 	& -16.76 	& -17.54 \\
O & C &  -37.84	& -53.88 	& -54.69 	&  -37.83 	& -53.73	& -54.89 \\
H & S &  -12.50 	& -16.08 	& -17.54 	&  -12.80 	& -16.44 	& -17.54 \\
H & C &  -41.70 	& -53.91 	& -54.69 	&  -40.08 	& -54.45 	& -54.89 \\
S & C &  494.9		& -54.49 	& -54.69 	&  498.9 	& -54.44 	& -54.89 \\
H & O &  -48.91 	& -55.08 	& -55.23 	&  -49.20 	& -54.84 	& -55.53 \\
S & O &  709.9 	& -54.07 	& -55.23 	&  709.9 	& -54.10 	& -55.53 \\
C & O &  -49.21 	& -55.00 	& -55.23 	& -49.17 	& -55.05	& -55.53 \\
S & H &  649.9 	& -56.09 	& -57.35 	&  650.3 	& -56.19 	& -57.62 \\
C & H &  -53.05 	& -57.19 	& -57.35 	& -53.15	& -57.17 	& -57.62 \\
C & S &  -15.45	& -17.43 	& -17.54 	& -15.47 	& -17.44 	& -17.54
\end{tabular}
\end{table}

\subsection{Results}
Table \ref{tab:hdis_tr} lists target risks for source, Target Robust and target classifiers for each possible pair of domains in the data set. In most cases, the TR classifier is quite close to the optimal target risk. Note that the source classifier performs terribly in some settings (with target risks in the positive hundreds), while it is does not differ much from the target classifier in others. 

\begin{table*}[htb] 
\caption{\label{tab:hdis_auc} AUC for all pairwise combinations of domains in heart disease data set (O='Ohio', H='Hungary', S='Switzerland' and C='California').}
%\centering
\setlength{\tabcolsep}{5pt}
\renewcommand{\arraystretch}{0.9}
\begin{tabular}{ r l | c c | c c c c c c }
$\mathcal{S}$ & $\mathcal{T}$ & \small S-LDA & \small S-QDA & \small TCA & \small KMM & \small RCSA & \small RBA & \small TR-LDA & \small TR-QDA \\
\midrule
O & H & 0.866 & 0.829 & 0.674 & 0.709 & 0.646 & 0.502 & 0.864 & 0.822 \\
O & S & 0.674 & 0.674 & 0.597 & 0.591 & 0.667 & 0.670 & 0.675 & 0.675 \\
O & C & 0.658 & 0.503 & 0.500 & 0.460 & 0.572 & 0.430 & 0.653 & 0.500 \\
H & S & 0.671 & 0.660 & 0.453 & 0.503 & 0.641 & 0.636 & 0.673 & 0.661 \\
H & C & 0.726 & 0.668 & 0.466 & 0.568 & 0.483 & 0.423 & 0.725 & 0.660 \\
S & C & 0.527 & 0.484 & 0.530 & 0.552 & 0.459 & 0.582 & 0.555 & 0.432 \\
H & O & 0.866 & 0.840 & 0.544 & 0.742 & 0.749 & 0.556 & 0.867 & 0.841 \\
S & O & 0.500 & 0.500 & 0.439 & 0.302 & 0.626 & 0.366 & 0.424 & 0.422 \\
C & O & 0.830 & 0.811 & 0.693 & 0.294 & 0.651 & 0.523 & 0.831 & 0.813 \\
S & H & 0.559 & 0.502 & 0.408 & 0.345 & 0.685 & 0.396 & 0.717 & 0.565 \\
C & H & 0.883 & 0.834 & 0.661 & 0.290 & 0.647 & 0.597 & 0.882 & 0.847 \\
C & S & 0.440 & 0.452 & 0.572 & 0.508 & 0.343 & 0.412 & 0.447 & 0.414 \\
\midrule
& avg & 0.683 & 0.647 & 0.545 & 0.489 & 0.597 & 0.508 & 0.693 & 0.638
\end{tabular}
\end{table*}

Table \ref{tab:hdis_auc} lists AUCs of different classifiers in the heart disease data set. Perhaps the most striking observation is that AUC's are sometimes below $0.5$; these scenarios represent domain shifts so large that source classifiers perform worse than chance in the target domain. Adaptive classifiers will not perform much better if their shift assumptions are violated. Our own TR classifiers are merely built to improve over their source counterpart: when the source classifier is poor to begin with, a "better'' performance may still be below chance. A few more things to note: firstly, TR-LDA generally outperforms TR-QDA, indicating that the additional flexibility of QDA does not outweigh the increase the complexity. Secondly, TR-LDA and TR-QDA are either performing similarly or better than S-LDA and S-QDA. Note that cases where the source classifiers perform well correspond to cases where the source classifier's target risk was small and close to that of the target classifier (compare to Table \ref{tab:hdis_tr}). Thirdly, RCSA and RBA do not always outperform KMM, indicating that robust weight estimation is not always beneficial. Fourthly, TCA's performance varies around chance level, which means that its assumption is likely violated.

%------------------------------
%------------------------------
%------------------------------
%------------------------------
%------------------------------

\section{Discussion}\label{disc}
As could be seen in the experimental results, an improvement in terms of the classifier's intrinsic loss does \emph{not} imply an improvement in AUC. This is due to the difference between optimizing a surrogate loss, here the negative log-likelihood, and evaluating the $0$/$1$-loss \cite{bartlett2006convexity,loog2016measuring}. They do not necessarily have the same minimizers. Note that the $0$/$1$-loss is not differentiable, and cannot be optimized over directly. We therefore argue that guarantees in terms of intrinsic losses are the most one can expect.

One advantage of our estimator is that we do not explicitly require source samples at training time. Our approach is therefore more memory-efficient than other domain-adaptive classifiers and more suited to privacy-sensitive supervised learning settings, such as federated learning.

\subsection{Related Work}\label{sect:rel}
Most methods for domain adaptation rely on an assumption of how the domains have shifted \cite{kouw2019review}. Examples of such assumptions include low joint-domain-error \cite{ben2010theory}, the existence of a domain-invariant subspace \cite{pan2011domain} and the assumption that only the covariates have shifted but not the posterior distributions \cite{huang2007correcting,cortes2014domain}. These assumptions may be implicit, for example domain-adversarial neural networks simultaneously minimize the divergence between the domains and train a source classifier which amounts to the low joint-domain error assumption \cite{ben2010theory}. Violations of assumptions mean adaptation could deteriorate performance. For example, Transfer Component Analysis assumes a domain-invariant latent representation where class separability is preserved \cite{pan2011domain}. When that assumption does not hold, mapping data onto transfer components will mix the class-conditional distributions and classification will become harder.

Research into robust domain adaptation tends to revolve around importance weight estimators for methods assuming covariate shift. Unfortunately, importance weight estimators may assign few samples large weights and many samples near-zero weights, greatly reducing effective sample size and producing pathological importance-weighted classifiers \cite{cortes2014domain}. Robust Covariate Shift Adjustment builds an importance-weighted classifier that is robust to poor importance weight estimates by first maximizing risk with respect to the importance-weights and subsequently minimizing with respect to classifier parameters \cite{wen2014robust}. However, it can perform worse than standard importance-weighted classifiers when it \emph{unnecessarily} considers worst-case weights. The Robust Bias-Aware classifier employs a similar mini-max strategy, but avoids accounting for worst-case importance weights. It attempts to match the statistics, specifically the moments, of the importance-weighted classifier's labelling of the target samples with the statistics of the source labels \cite{liu2014robust}. 
This favours more stable importance-weighted classifiers, but the RBA classifier loses predictive power in areas of feature space where the source distribution's support is limited.

Similar research concerning improvement guarantees has been carried out: Maximum Contrastive Pessimistic Likelihood estimation is a worst-case approach to semi-supervised learning that ensures complete robustness to the labelling of the unlabelled samples \cite{loog2016contrastive}. It also comes with performance guarantees in terms of the objective that the classifier actually optimizes, such as log-likelihood, hinge loss, logistic loss, etc. \cite{loog2016measuring}.

\section{Conclusion}\label{conc}
We have designed a risk minimization formulation for a domain-adaptive classifier whose performance, in terms of empirical target risk, is always at least as good as that of the non-adaptive source classifier. Furthermore, for the discriminant analysis case, its risk is always strictly smaller. An experiment on data gathered under a geographical bias supports the claim empirically and shows competitive performance compared to other robust domain-adaptive classifiers.

\bibliographystyle{splncs04}
\bibliography{kouw_ssspr2020}

\end{document}